\documentclass{article} % For LaTeX2e
\pdfoutput=1 

\usepackage{iclr2019_conference,times}

\usepackage{caption}
\usepackage{url}
\usepackage{xcolor}
\usepackage{tabularx}

\usepackage{pgfplotstable} %tikz plots data
\usepackage{pgfplots}
\pgfplotsset{compat=1.14}

\usepackage{mathtools}
\usepackage{amsfonts}
\usepackage{mathtools}
\usepackage[%
    mode=text,
    group-separator={,},
    group-minimum-digits = 4,
]{siunitx}

\usepackage{longtable}
\usepackage{booktabs}
\usepackage[colorlinks,allcolors=black]{hyperref}
\hypersetup{
  pdfinfo={
    Title={wav2vec: Unsupervised Pre-Training for Speech Recognition},
    Author={Steffen Schneider, Alexei Baevski, Ronan Collobert, Michael Auli},
  }
}

\graphicspath{{./figs/} {./figs/figures/}}

\makeatletter
\renewcommand*{\@fnsymbol}[1]{\ensuremath{\ifcase#1\or * \or \dagger\or \ddagger\or
    \mathsection\or \mathparagraph\or \|\or **\or \dagger\dagger
    \or \ddagger\ddagger \else\@ctrerr\fi}}
\makeatother

\newcommand{\wav}{wav2vec}
\newcommand{\wavl}{wav2vec large}
\newcommand{\libri}{Librispeech}

\newcommand{\wavtoletter}{wav2letter\texttt{+\!+}}
\DeclareMathOperator*{\Exp}{\mathbb{E}}

\newcommand{\Inp}{\mathcal{X}}
\newcommand{\Feat}{\mathcal{Z}}
\newcommand{\Context}{\mathcal{C}}

\newcommand{\x}{\mathbf{x}}
\newcommand{\y}{\mathbf{y}}
\newcommand{\z}{\mathbf{z}}
\newcommand{\tz}{\mathbf{\tilde{z}}}
\newcommand{\cc}{\mathbf{c}}

\usepackage{tikz}
\usepackage{pgfplots}
\pgfplotsset{compat=1.13}
\pgfplotsset{
    table/search path={figs/data},
}

\usepackage{xfrac}
\usetikzlibrary{arrows}

\usetikzlibrary{math}
\usetikzlibrary{shapes,arrows}
\usetikzlibrary{positioning}
\usetikzlibrary{decorations,patterns}
\usetikzlibrary{calc,fit,matrix,chains,scopes}

\newcommand{\lw}{1pt}

\definecolor{FAIRlg}{HTML}{F7F4F0}
\definecolor{FAIRlb}{HTML}{4267B6}
\definecolor{FAIRdb}{HTML}{003462}

\definecolor{FAIRaclg}{HTML}{F4F4F4}
\definecolor{FAIRaclb}{HTML}{EDF0F6}
\definecolor{FAIRdark}{HTML}{162643}

\definecolor{accent1}{HTML}{4267B6}
\definecolor{accent2}{HTML}{003462}
\definecolor{accent3}{HTML}{162643}

\definecolor{myblue}{RGB}{31, 119, 180}
\definecolor{myred}{RGB}{44, 160, 44}
\definecolor{mygreen}{RGB}{255, 127, 14}
\definecolor{myviolet}{RGB}{140, 86, 75}

\tikzset{%
    block/.style        = {line width=\lw, rectangle, draw=black,dashed, minimum width=\dist, minimum height=.33*\dist},
    representation/.style        = {line width=\lw, rectangle, draw=white!0,fill=mygreen, minimum width=.05*\dist, minimum height=.3\dist},
    filled/.style        = {line width=\lw, rectangle, draw=white!0, rounded corners, minimum width=.15*\dist, minimum height=.05*\dist},
    line/.style         = {draw=gray, -latex, line width=.75*\lw,rounded corners},
    lineplain/.style    = {draw=gray, -, line width=.75*\lw, rounded corners},
    mathop/.style       = {draw, circle, fill=gray!20},
    branch/.style       = {fill,shape=circle,minimum size=\dist,inner sep=0pt},
    encoder/.style      = {line width=\lw, trapezium, trapezium angle=80, rotate=-90, minimum width=.55*\dist,  draw, trapezium stretches=true, fill=myblue,draw=white!0,minimum height=.15*\dist},
    classifier/.style      = {line width=\lw, trapezium, trapezium angle=80, rotate=-90, minimum width=.35*\dist, minimum height=.15*\dist, draw, trapezium stretches=true, draw=white},
    decoder/.style      = {line width=\lw, trapezium, trapezium angle=80, rotate=90,  minimum width=.55*\dist, minimum height=.15*\dist, draw=white, trapezium stretches=true},
    mult/.style={path picture={
      \draw[black]
    (path picture bounding box.south east) -- (path picture bounding box.north west) (path picture bounding box.south west) -- (path picture bounding box.north east);
    }},
    add/.style={path picture={
      \draw[black]
    (path picture bounding box.south) -- (path picture bounding box.north) (path picture bounding box.west) -- (path picture bounding box.east);
    }}
}

\makeatletter
\pgfplotsset{
  tufte axes/.style =
    {
      after end axis/.code =
        {
          \draw ({rel axis cs:0,0} -| {axis cs:\pgfplots@data@xmin,0})
            -- ({rel axis cs:0,0}  -| {axis cs:\pgfplots@data@xmax,0});
          \draw ({rel axis cs:0,0} |- {axis cs:0,\pgfplots@data@ymin})
            -- ({rel axis cs:0,0}  |-{axis cs:0,\pgfplots@data@ymax});
        },
      axis line style = {draw = none},
      tick align      = outside,
      tick pos        = left
    }
}
\makeatother

\newcommand{\dist}{10em}

\title{\wav: Unsupervised Pre-training for \\Speech Recognition}

\author{Steffen Schneider, Alexei Baevski, Ronan Collobert, Michael Auli \\
Facebook AI Research \\
}

\iclrfinalcopy

\usepackage{ifthen}
\newcommand{\fullwsjresult}{\hyperref[tbl:wsj-results]{{\SI{2.43}{\percent}}}}
\newcommand{\lowresourceresult}{\hyperref[fig:low-resource]{{\SI{36}{\percent}}}}

\begin{document}

\maketitle

\begin{abstract}
We explore unsupervised pre-training for speech recognition by learning representations of raw audio.
wav2vec is trained on large amounts of unlabeled audio data and the resulting representations are then used to improve acoustic model training.
We pre-train a simple multi-layer convolutional neural network optimized via a noise contrastive binary classification task.
Our experiments on WSJ reduce WER of a strong character-based log-mel filterbank baseline by up to \lowresourceresult{} when only a few hours of transcribed data is available.
Our approach achieves \fullwsjresult{} WER on the nov92 test set. 
This outperforms Deep Speech 2, the best reported character-based system in the literature while using two orders of magnitude less labeled training data.\footnote{The code is available as part of fairseq (\url{https://github.com/pytorch/fairseq}).}
\end{abstract}

\section{Introduction}

Current state of the art models for speech recognition require large amounts of transcribed audio data to attain good performance~\citep{amodei2016deepspeech}.
Recently, pre-training of neural networks has emerged as an effective technique for settings where labeled data is scarce.
The key idea is to learn general representations in a setup where substantial amounts of labeled or unlabeled data is available and to leverage the learned representations to improve performance on a downstream task for which the amount of data is limited.
This is particularly interesting for tasks where substantial effort is required to obtain labeled data, such as speech recognition.

In computer vision, representations for ImageNet~\citep{deng2009imagenet} and COCO~\citep{lin2014microsoft} have proven to be useful to initialize models for tasks such as image captioning~\citep{vinyals2016show} or pose estimation~\citep{pavllo2019cvpr}. 
Unsupervised pre-training for computer vision has also shown promise~\citep{doersch2015unsup,oord2019cpc}.
In natural language processing (NLP), unsupervised pre-training of language models~\citep{devlin2018bert,radford2018unsup,baevski2019cloze} improved many tasks such as text classification, phrase structure parsing and machine translation~\citep{edunov2019pretrain,lample2019cross}.
In speech processing, pre-training has focused on emotion recogniton~\citep{lian2018emotion}, speaker identification~\citep{ravanelli2018mutual}, phoneme discrimination~\citep{synnaeve2016coherence,oord2018cpc} as well as transferring ASR representations from one language to another~\citep{kunze2017transfer}.
There has been work on unsupervised learning for speech but the resulting representations have not been applied to improve supervised speech recognition~\citep{synnaeve2016temporal,kamper2017seg,chung2018unsup,chen2018unsup,chorowski2019unsup}.

In this paper, we apply unsupervised pre-training to improve supervised speech recognition.
This enables exploiting unlabeled audio data which is much easier to collect than labeled data.
Our model, \wav{}, is a convolutional neural network that takes raw audio as input and computes a general representation that can be input to a speech recognition system.
The objective is a contrastive loss that requires distinguishing a true future audio sample from negatives~\citep{collobert2011jmlr,mikolov2013word2vec,oord2018cpc}.
Different to previous work~\citep{oord2018cpc}, we move beyond frame-wise phoneme classification and apply the learned representations to improve strong supervised ASR systems. 
\wav{} relies on a fully convolutional architecture which can be easily parallelized over time on modern hardware compared to recurrent models used in previous work~(\textsection\ref{sec:approach}).

Experimental results on the WSJ benchmark demonstrate that pre-trained representations estimated on about \num{1000} hours of unlabeled speech can substantially improve a character-based ASR system and outperform the best character-based result in the literature, Deep Speech 2, improving WER from \SI{3.1}{\percent} to \fullwsjresult{}.
On TIMIT, pre-training enables us to match the best reported result in the literature.
In a simulated low-resource setup with only eight hours of transcribed audio data, \wav{} reduces WER by up to \lowresourceresult{} compared to a baseline model that relies on labeled data only (\textsection\ref{sec:setup}, \textsection\ref{sec:results}).

\begin{figure}[t]
\input{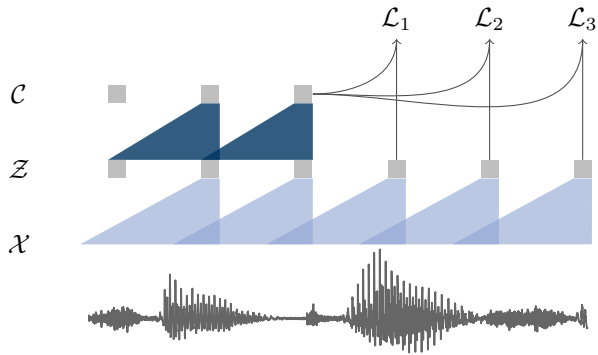}

\caption{Illustration of pre-training from audio data $\mathcal{X}$ which is encoded with two convolutional neural networks that are stacked on top of each other. 
The model is optimized to solve a next time step prediction task.
}
\label{fig:model}
\end{figure}

\section{Pre-training Approach}
\label{sec:approach}

Given an audio signal as input, we optimize our model (\textsection\ref{sec:model}) to predict future samples from a given signal context.
A common problem with these approaches is the requirement to accurately model the data distribution $p(\x)$, which is challenging. 
We avoid this problem by first encoding raw speech samples $\x$ into a feature representation $\z$ at a lower temporal frequency and then implicitly model a density ratio $p(\z_{i+k} | \z_i \dots \z_{i-r})/p(\z_{i+k})$ similar to~\citet{oord2018cpc}.

\subsection{Model}
\label{sec:model}

Our model takes raw audio signal as input and then applies two networks. 
The encoder network embeds the audio signal in a latent space and the context network combines multiple time-steps of the encoder to obtain contextualized representations (Figure~\ref{fig:model}). 
Both networks are then used to compute the objective function~(\textsection\ref{sec:objective}).

Given raw audio samples $\x_i \in \Inp$, we apply the \emph{encoder} network $f: \Inp \mapsto \Feat$ parameterized as a five-layer convolutional network~\citep{oord2018cpc}.
Alternatively, one could use other architectures such as the trainable frontend of ~\citet{zeghidour2018filters}.
The encoder layers have kernel sizes $(10,8,4,4,4)$ and strides $(5,4,2,2,2)$.
The output of the encoder is a low frequency feature representation $\z_i \in \Feat$ which encodes about \SI{30}{\ms} of \SI{16}{\kHz} of audio and the striding results in representations $\z_i$ every 10ms.

Next, we apply the \emph{context} network $g: \Feat \mapsto \Context$ to the output of the encoder network to mix multiple latent representations $\z_i \dots \z_{i-v}$ into a single contextualized tensor $\cc_{i} = g( \z_i \dots \z_{i-v} )$ for a receptive field size $v$.
The context network has nine layers with kernel size three and stride one.
The total receptive field of the context network is about \SI{210}{\ms}.

The layers in both the encoder and context networks consist of a causal convolution with 512 channels, a group normalization layer and a ReLU nonlinearity.
We normalize both across the feature and temporal dimension for each sample which is equivalent to group normalization with a single normalization group~\citep{wu2018gn}.
We found it important to choose a normalization scheme that is invariant to the scaling and the offset of the input.
This choice resulted in representations that generalize well across datasets.

For training on larger datasets, we also consider a model variant (``\wav{} large'') with increased capacity, using two additional linear transformations in the encoder and a considerably larger context network comprised of twelve layers with increasing kernel sizes $(2,3,\dots,13)$.
We found it important to introduce skip connections in the aggregator to help convergence in this case.
The total receptive field in the last context network layer is hereby increased to about \SI{810}{\ms}.

\subsection{Objective}
\label{sec:objective}

We train the model to distinguish a sample $\z_{i+k}$ that is k steps in the future from distractor samples $\tz$ drawn from a proposal distribution $p_n$, by minimizing the contrastive loss for each step $k=1,\dots,K$:
\begin{equation}
  \mathcal{L}_k = - \sum_{i=1}^{T-k} \Big(
  \log \sigma(\z_{i+k}^\top h_k(\cc_i)) +
  \lambda \Exp_{\mathclap{\tz \sim p_n}}\ [ \log \sigma(-\tz^\top h_k(\cc_i)) ]\ \Big)\,
  \label{eq:objective}
\end{equation}
where we denote the sigmoid $\sigma(x) = 1/(1+\exp(-x))$, and where $\sigma(\z_{i+k}^\top h_k(\cc_i))$ is the probability of $\z_{i+k}$ being the true sample. We consider a step-specific affine transformation $h_k(\cc_i) = W_k \cc_i + \mathbf{b}_k$ for each step $k$, that is applied to $\cc_i$~\citep{oord2018cpc}.
We optimize the loss $\mathcal{L} = \sum_{k=1}^K \mathcal{L}_k$, summing~(\ref{eq:objective}) over different step sizes.
In practice, we approximate the expectation by sampling ten negatives examples by uniformly choosing distractors from each audio sequence, i.e., $p_n(\z) = \frac{1}{T}$, where $T$ is the sequence length and we set $\lambda$ to the number of negatives.\footnote{Similar to~\citet{oord2018cpc}, we found that sampling negatives from different sequences and speakers yields inferior results.}

After training, we input the representations $\cc_i$  produced by the context network to the acoustic model instead of log-mel filterbank features.

\section{Experimental Setup}
\label{sec:setup}

\subsection{Data}
\label{sec:data}

We consider the following corpora:
For phoneme recognition on TIMIT~\citep{garofolo1993timit} we use the standard train, dev and test split where the training data contains just over three hours of audio data.
Wall Street Journal (WSJ; \citet{garofolo1993wsj,woodland1994large}) comprises about 81 hours of transcribed audio data.
We train on si284, validate on nov93dev and test on nov92.
Librispeech~\citep{panayotov2015librispeech} contains a total of 960 hours of clean and noisy speech for training.
For pre-training, we use either the full 81 hours of the WSJ corpus, an 80 hour subset of clean Librispeech, the full 960 hour Librispeech training set or a combination of all of them.

To train the baseline acoustic model we compute 80 log-mel filterbank coefficients for a \SI{25}{\ms} sliding window with stride \SI{10}{\ms}.
Final models are evaluated in terms of both word error rate (WER) and letter error rate (LER).

\subsection{Acoustic Models}

We use the \wavtoletter{} toolkit for training and evaluation of acoustic models~\citep{pratap2018w2l}.
% TIMIT
For the TIMIT task, we follow the character-based \wavtoletter{} setup of~\citet{zeghidour2018filters} which uses seven consecutive blocks of convolutions (kernel size \num{5} with \num{1000} channels), followed by a PReLU nonlinearity and a dropout rate of \num{0.7}.
The final representation is projected to a 39-dimensional phoneme probability.
The model is trained using the Auto Segmentation Criterion (ASG; Collobert et al., 2016)\nocite{collobert2016wav2letter}) using SGD with momentum.

% WSJ
Our baseline for the WSJ benchmark is the \wavtoletter{} setup described by~\citet{collobert2018diffbeam} which is a 17 layer model with gated convolutions~\citep{dauphin2016convlm}.
The model predicts probabilities for 31 graphemes, including the standard English alphabet, the apostrophe and period, two repetition characters (e.g. the word \texttt{ann} is transcribed as \texttt{an1}), and a silence token (\texttt{|}) used as word boundary.

All acoustic models are trained on 8 \textsc{Nvidia} V100 GPUs using the distributed training implementations of fairseq and \wavtoletter{}.
When training acoustic models on WSJ, we use plain SGD with learning rate 5.6 as well as gradient clipping~\citep{collobert2018diffbeam} and train for 1,000 epochs with a total batch size of 64 audio sequences.
We use early stopping and choose models based on validation WER after evaluating checkpoints with a 4-gram language model.
For TIMIT we use learning rate 0.12, momentum 0.9 and we train for \num{1000} epochs on 8 GPUs with a batch size of 16 audio sequences.

% !TEX root = ../paper.tex

\begin{table}[t]
\centering
\begin{tabular}{p{35mm}p{18mm}>{\raggedleft\arraybackslash}p{14mm}p{8mm}p{8mm}p{8mm}p{8mm}}
\toprule
&&& \multicolumn{2}{c}{nov93dev} & \multicolumn{2}{c}{nov92} \\
&&&   LER &    WER &      LER &    WER \\
\midrule
% Character-based models \\
\multicolumn{3}{l}{%
Deep Speech 2 (12K h labeled speech; Amodei et al., 2016)\nocite{amodei2016deepspeech} 
}
    & -
    & 4.42
    & -
    & 3.1 \\
\multicolumn{3}{l}{%
Trainable frontend~\citep{zeghidour2018filters} 
}
    & - 
    & 6.8 
    & - 
    & 3.5 \\
\multicolumn{3}{l}{%
Lattice-free MMI~\citep{hadian2018interspeech}
}
   & -
   & 5.66$^\dagger$ 
   & - 
   & 2.8$^\dagger$ \\
\multicolumn{3}{l}{%
Supervised transfer-learning~\citep{ghahremani2017asru} 
}
  & - 
  & 4.99$^\dagger$ 
  & - 
  & 2.53$^\dagger$ \\
\midrule
\multicolumn{2}{l}{\textsc{4-gram LM} \citep{heafield2013kenlm}} & \\
Baseline %log-mel filterbanks 
    & -- & --
    & 3.32
    & 8.57
    & 2.19
    & 5.64 \\
\wav{} & \libri{} & \SI{80}{\hour} 
    & 3.71 
    & 9.11 
    & 2.17 
    & 5.55 \\ 
\wav{} & \libri{} & \SI{960}{\hour} 
    & 2.85
    & 7.40
    & 1.76
    & 4.57 \\
\wav{} & Libri + WSJ & \SI{1041}{\hour}
    & 2.91
    & 7.59
    & 1.67
    & 4.61 \\
\wavl{} & \libri{} & \SI{960}{\hour}
    & 2.73
    & 6.96
    & 1.57
    & 4.32 \\
\midrule
\multicolumn{2}{l}{\textsc{Word ConvLM} \citep{zeghidour2018w2l}} & \\
Baseline 
    & -- & --
    & 2.57 & 6.27 & 1.51 & 3.60 \\
\wav{} & \libri{} & \SI{960}{\hour}
    & 2.22 &	5.39 &	1.25 &	2.87 \\
\wavl{} & \libri{} & \SI{960}{\hour}
    & 2.13
    & 5.16
    & 1.02
    & 2.53 \\
    \midrule
\multicolumn{2}{l}{\textsc{Char ConvLM} \citep{likhomanenko2019convlm}} & \\
Baseline %log-mel filterbanks 
    & -- & -- & 2.77 & 6.67 &	1.53	& 3.46 \\
\wav{} & \libri{} & \SI{960}{\hour}
    & 2.14 
    & 5.31
    & 1.15
    & 2.78 \\
\wavl{} & \libri{} & \SI{960}{\hour}
    & 2.11	
    & 5.10	
    & 0.99	
    & 2.43 \\
\bottomrule
\end{tabular}
\caption{
Replacing log-mel filterbanks (Baseline) by pre-trained embeddings improves WSJ performance on test (nov92) and validation (nov93dev) in terms of both LER and WER.
We evaluate pre-training on the acoustic data of part of clean and full Librispeech as well as the combination of all of them.
$^\dagger$\,indicates results with phoneme-based models.
}
\label{tbl:wsj-results}
\end{table}

\subsection{Decoding}

For decoding the emissions from the acoustic model we use a lexicon as well as a separate language model trained on the WSJ language modeling data only. 
We consider a 4-gram KenLM language model~\citep{heafield2013kenlm}, a word-based convolutional language model~\citep{collobert2018diffbeam}, and a character based convolutional language model~\citep{likhomanenko2019convlm}.
We decode the word sequence $\y$ from the output of the context network $\cc$ or log-mel filterbanks using the beam search decoder of~\citet{collobert2018diffbeam} by maximizing
\begin{equation}
  \max_{\y} f_\text{AM}(\y | \cc )
              + \alpha \log p_\text{LM} (\y)
              + \beta | \y |
              - \gamma \sum_{i=1}^T [\pi_i=`|`]
\end{equation}
where $f_\text{AM}$ is the acoustic model, $p_\text{LM}$ is the language model, $\pi = \pi_1, ..., \pi_L$ are the characters of $\y$. Hyper-parameters $\alpha$, $\beta$ and $\gamma$ are weights for the language model, the word penalty, and the silence penalty.

For decoding WSJ, we tune the hyperparameters $\alpha$, $\beta$ and $\gamma$ using a random search.
Finally, we decode the emissions from the acoustic model with the best parameter setting for $\alpha$, $\beta$ and $\gamma$.
We use a beam size of \num{4000} and beam score threshold of \num{250} for the word based language models, and a beam size of \num{1500} with beam score threshold \num{40} for the character based language model.

\subsection{Pre-training Models}
\label{sec:setup-pretrain}

The pre-training models are implemented in PyTorch in the fairseq toolkit~\citep{ott2019fairseq}.
We optimize them with Adam~\citep{kingma2015adam} and a cosine learning rate schedule~\citep{loshchilov2016cosine} annealed over 40\si{\kilo} update steps for both WSJ and the clean Librispeech training datasets or over 400\si{\kilo} steps for full Librispeech.
We start with a learning rate of \num[exponent-product = \times]{1e-7}, and then gradually warm it up for 500 updates up to \num[exponent-product = \times]{5e-3} and then decay it following the cosine curve up to \num[exponent-product = \times]{1e-6}.
To compute the objective, we sample ten negatives for each of the $K=12$ tasks.

We train the first wav2vec variant on 8 GPUs and put audio sequences amounting up to 1.5M frames on each GPU.
Sequences are grouped by length and we crop each to a maximum size of {150}\si{\kilo} frames, or the length of the shortest sequence in the batch, whichever is smaller. 
Cropping removes speech signal from either the beginning or the end of the sequence and we randomly decide the cropping offsets for each sample; we re-sample every epoch.
This is a form of data augmentation but also ensures equal length of all sequences on a GPU and removes on average \SI{25}{\percent} of the training data.
After cropping, the total effective batch size across GPUs is about 556 seconds of speech. %signal (for a variable number of audio sequences).
For the large model variant, we train on 16 GPUs, doubling the effective batch size.

\section{Results}
\label{sec:results}

Different to~\citet{oord2018cpc}, we evaluate the pre-trained representations directly on downstream speech recognition tasks. 
We measure speech recognition performance on the WSJ benchmark and simulate various low resource setups (\textsection\ref{sec:res-wsj}). 
We also evaluate on the TIMIT phoneme recognition task (\textsection\ref{sec:res-timit}) and ablate various modeling choices (\textsection\ref{sec:ablations}).

\subsection{Pre-training for the WSJ benchmark}
\label{sec:res-wsj}

\pgfkeys{/pgfplots/tuftelike/.style={
  thick,
  tick style={major tick length=3pt, thick, black},
  axis x line*=bottom,
  axis line shift = 10pt,
  tick align      = outside,
  tick pos        = left,
  xlabel shift=0pt,
  axis y line*=left,
  ylabel shift=0pt}
}
  
\pgfmathsetlengthmacro\MajorTickLength{
  \pgfkeysvalueof{/pgfplots/major tick length} * 4
}

\pgfkeys{%
    /pgfplots/plotone/.style={
        dashed, mark=*, mark options={solid,fill=blue}, draw=blue
    },
    /pgfplots/plottwo/.style={
        dashdotted, mark=triangle*, mark options={solid,fill=red}, draw=red
    },
    /pgfplots/plotthree/.style={
        mark=square*, mark options={fill=brown, scale=0.75}, draw=brown
    }
}

\begin{figure*}[t]
\begin{center}
\begin{tikzpicture}
\begin{axis}[
tuftelike,
y tick label style={
    /pgf/number format/.cd,
        fixed,
        fixed zerofill,
        precision=1,
    /tikz/.cd
},
width=0.4\columnwidth,
height=0.3\textwidth,
xmode=log,
xlabel=hours of labeled si284 (training set),
ylabel=nov93dev WER,
xtick={7.59,16,33,81.4},
xticklabels={8,16,33,81},
ymin=7.39616,
ymax=23.0993,
ytick={%
    7.39616,
    13.7965,
    18.05,
    23.0993
},
xmin=7.59,
xmax=82,
legend pos=north east,
legend style={draw = none, at={(0.99,0.99)}}
]
\addplot[plotone] table [y=base, x=hrs]{figs/data/low-resource-nov93dev-wer.dat};
\addplot[plottwo] table [y=wsj, x=hrs]{figs/data/low-resource-nov93dev-wer.dat};
\addplot[plotthree] table [y=libri960, x=hrs]{figs/data/low-resource-nov93dev-wer.dat};

\draw   (9.3, 23.0993)
     to (9.9, 23.0993) 
     to (9.9, 13.7965) 
     to (9.3, 13.7965);
\node (annotation) at (14, 22) {\footnotesize \SI{-40}{\percent}};
\end{axis}

\begin{axis}[
at={(0.45\columnwidth, 0)},
tuftelike,
y tick label style={
    /pgf/number format/.cd,
        fixed,
        fixed zerofill,
        precision=1,
    /tikz/.cd
},
width=0.4\columnwidth,
height=0.3\textwidth,
axis line shift = 10pt,
xmode=log,
xlabel=hours of labeled si284  (training set),
ylabel=nov92 WER,
xtick={7.59,16,33,81.4},
xticklabels={8,16,33,81},
ymax=15.2401,
ymin=4.57204,
ytick={%
    4.57204,
    9.69343,
    12.60,
    15.2401
},
xmin=7.59,
xmax=82,
legend pos=north east,
legend style={draw = none, at={(1.2,1.05)}}
]
\addplot [plotone] table [y=base, x=hrs]{figs/data/low-resource-nov92-wer.dat};
\addlegendentry{Baseline}
\addplot [plottwo] table [y=wsj, x=hrs]{figs/data/low-resource-nov92-wer.dat};
\addlegendentry{\wav{} WSJ}
\addplot [plotthree] table [y=libri960, x=hrs]{figs/data/low-resource-nov92-wer.dat};
\addlegendentry{\wav{} Libri}

\draw   (9.3,15.2401)
     to (9.9,15.2401) 
     to (9.9, 9.69343) 
     to (9.3, 9.69343);

\coordinate (thisannot) at (14, 0);
\node at (annotation -| thisannot) {\footnotesize \SI{-36}{\percent}};

\end{axis}

\end{tikzpicture}
\caption{
Pre-training substanstially improves WER in simulated low-resource setups on the audio data of WSJ compared to \wavtoletter{} with log-mel filterbanks features (Baseline). 
Pre-training on the audio data of the full 960\,h Librispeech dataset (\wav{} Libri) performs better than pre-training on the 81\,h WSJ dataset (\wav{} WSJ).
}
\label{fig:low-resource}
\end{center}
\end{figure*}
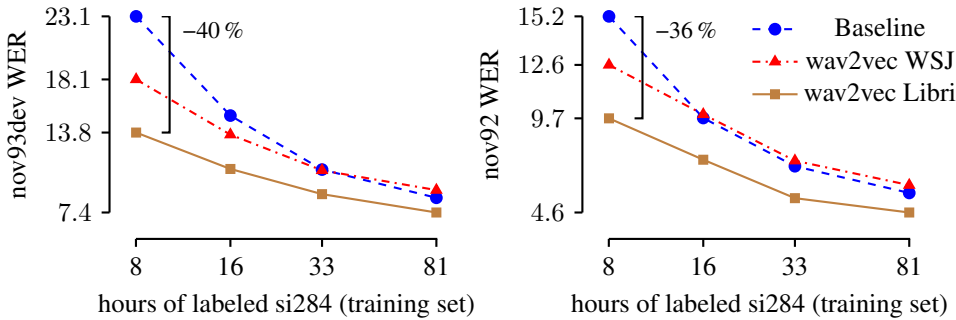

We consider pre-training on the audio data (without labels) of WSJ, part of clean Librispeech (about \SI{80}{\hour}) and full Librispeech as well as a combination of all datasets (\textsection\ref{sec:data}).
For the pre-training experiments we feed the output of the context network to the acoustic model, instead of log-mel filterbank features.

Table~\ref{tbl:wsj-results} shows that pre-training on more data leads to better accuracy on the WSJ benchmark.
Pre-trained representations can substantially improve performance over our character-based baseline which is trained on log-mel filterbank features.
This shows that pre-training on unlabeled audio data can improve over the best character-based approach, Deep Speech 2~\citep{amodei2016deepspeech}, by 0.67 WER on nov92.
In comparison to~\citet{hadian2018interspeech}, \wav{} performs as well as their phoneme-based model and \wavl{} outperforms it by 0.37 WER.
The phoneme-based approach of~\citet{ghahremani2017asru} pre-trains on the labeled version of Librispeech and then fine-tunes on WSJ.
\wavl{} still outperforms~\citet{ghahremani2017asru} despite a weaker baseline model and not using Librispeech transcriptions.

What is the impact of pre-trained representations with less transcribed data?
In order to get a better understanding of this, we train acoustic models with different amounts of labeled training data and measure accuracy with and without pre-trained representations (log-mel filterbanks).
The pre-trained representations are trained on the full Librispeech corpus and we measure accuracy in terms of WER when decoding with a 4-gram language model.
Figure~\ref{fig:low-resource} shows that pre-training reduces WER by \lowresourceresult{} on nov92 when only about eight hours of transcribed data is available.
Pre-training only on the audio data of WSJ (\wav{} WSJ) performs worse compared to the much larger Librispeech (\wav{} Libri).
This further confirms that pre-training on more data is important to good performance.
Similar to \citet{oord2019cpc}, we noticed that fine-tuning the embedding network does not meaningfully improve performance while substantially increasing the acoustic model training time.

\subsection{Pre-training for TIMIT}
\label{sec:res-timit}

% !TEX root = ../paper.tex

\begin{table}[t]
\centering
\begin{tabular}{lll}
\toprule
{} &   dev &  test \\
\midrule
CNN + TD-filterbanks~\citep{zeghidour2018filters} & 15.6 & 18.0 \\
Li-GRU + MFCC~\citep{ravanelli2018light} & -- & 16.7 $\pm$ 0.26 \\
Li-GRU + FBANK~\citep{ravanelli2018light} & -- & 15.8 $\pm$ 0.10 \\
Li-GRU + fMLLR~\citep{ravanelli2018light} & -- & 14.9 $\pm$ 0.27 \\
\midrule
Baseline & 16.9 $\pm$ 0.15 & 17.6 $\pm$ 0.11  \\
\wav{} (\libri{} 80h) & 15.5 $\pm$ 0.03 & 17.6 $\pm$ 0.12 \\
\wav{} (\libri{} 960h) & 13.6 $\pm$ 0.20 & 15.6 $\pm$ 0.23 \\
\wav{} (\libri{} + WSJ) & \textbf{12.9 $\pm$ 0.18} & \textbf{14.7 $\pm$ 0.42} \\
\bottomrule
\end{tabular}
\caption{%
Results for phoneme recognition on TIMIT in terms of PER.
All our models use the CNN-8L-PReLU-do0.7 architecture~\citep{zeghidour2018filters}.
}
\label{tbl:timit-results}
\end{table}

On the TIMIT task we use a 7-layer \wavtoletter{} model with high dropout (\textsection\ref{sec:setup}; \citet{synnaeve2016temporal}).
Table~\ref{tbl:timit-results} shows that wav2vec pre-training on Librispeech and WSJ audio data can lead to results matching the state of the art. 
Accuracy steadily increases with more data for pre-training and the best accuracy is achieved with the largest amount of data for pre-training.

\subsection{Ablations}
\label{sec:ablations}

%!TEX root = ../paper.tex

\begin{table}[t]
\centering

\begin{tabular}{lrr}
\toprule
negatives &  dev PER & train time (h)\\
\midrule
1 & 16.3 & 6.1 \\
2 & 15.8 & 6.3 \\
5 &	15.9 & 8.2 \\
10 & 15.5 & 10.5 \\
20 & 15.7 & 15.3 \\
\bottomrule
\end{tabular}
\caption{%
Effect of different number of negative samples during pre-training for TIMIT on the development set.
}
\label{tbl:negatives}
\end{table}
%!TEX root = ../paper.tex

\begin{table}[t]
\centering
\begin{minipage}{0.4\textwidth}
\centering
    \begin{tabular}[t]{lr}
    \toprule
    Crop size & dev PER \\
    \midrule 
    None (Avg. 207\si{\kilo}) & 16.3 \\
    100\si{\kilo} & 16.1 \\
    150\si{\kilo} & \textbf{15.5} \\
    200\si{\kilo} & 16.0 \\
    \bottomrule
    \end{tabular}
    \caption{%
Effect of different crop sizes (cf. Table~\ref{tbl:negatives}).
}
\label{tbl:cropping}
\end{minipage}\qquad
\begin{minipage}{0.4\textwidth}
\centering
    \begin{tabular}[t]{lr}
    \toprule
    \# Tasks & dev PER \\
    \midrule 
    \\
    8 & 15.9 \\
    12 & \textbf{15.5} \\
    16 & 15.5 \\
    \bottomrule
    \end{tabular}
\caption{%
Effect of different number of tasks $K$ (cf. Table~\ref{tbl:negatives}).
}
\label{tbl:steps}
\end{minipage}
\end{table}

In this section we analyze some of the design choices we made for \wav{}.
We pre-train on the 80 hour subset of clean Librispeech and evaluate on TIMIT.
Table~\ref{tbl:negatives} shows that increasing the number of negative samples only helps up to ten samples.
Thereafter, performance plateaus while training time increases.
We suspect that this is because the training signal from the positive samples decreases as the number of negative samples increases.
In this experiment, everything is kept equal except for the number of negative samples.

Next, we analyze the effect of data augmentation through cropping audio sequences (\textsection\ref{sec:setup-pretrain}).
When creating batches we crop sequences to a pre-defined maximum length.
Table~\ref{tbl:cropping} shows that a crop size of 150\si{\kilo} frames results in the best performance. 
Not restricting the maximum length (None) gives an average sequence length of about 207\si{\kilo} frames and results in the worst accuracy.
This is most likely because the setting provides the least amount of data augmentation.

Table~\ref{tbl:steps} also shows that predicting more than 12 steps ahead in the future does not result in better performance and increasing the number of steps increases training time.

\section{Conclusions}

We introduce \wav{}, the first application of unsupervised pre-training to speech recognition with a fully convolutional model.
Our approach achieves \fullwsjresult{} WER on the test set of WSJ, a result that outperforms the next best known character-based speech recognition model in the literature~\citep{amodei2016deepspeech} while using two orders of magnitude less transcribed training data.
We show that more data for pre-training improves performance and that this approach not only improves resource-poor setups, but also settings where all WSJ training data is used.
In future work, we will investigate different architectures which is likely to further improve performance.

\section*{Acknowledgements}

We thank the Speech team at FAIR, especially Jacob Kahn, Vineel Pratap and Qiantong Xu for help with \wavtoletter{} experiments, and Tatiana Likhomanenko for providing convolutional language models for our experiments.

\bibliographystyle{iclr2019_conference}

\small
\bibliography{master}

\begin{thebibliography}{43}
\providecommand{\natexlab}[1]{#1}
\providecommand{\url}[1]{\texttt{#1}}
\expandafter\ifx\csname urlstyle\endcsname\relax
  \providecommand{\doi}[1]{doi: #1}\else
  \providecommand{\doi}{doi: \begingroup \urlstyle{rm}\Url}\fi

\bibitem[Amodei et~al.(2016)Amodei, Ananthanarayanan, Anubhai, Bai, Battenberg,
  Case, Casper, Catanzaro, Cheng, Chen, et~al.]{amodei2016deepspeech}
Dario Amodei, Sundaram Ananthanarayanan, Rishita Anubhai, Jingliang Bai, Eric
  Battenberg, Carl Case, Jared Casper, Bryan Catanzaro, Qiang Cheng, Guoliang
  Chen, et~al.
\newblock Deep speech 2: End-to-end speech recognition in english and mandarin.
\newblock In \emph{Proc. of ICML}, 2016.

\bibitem[Baevski et~al.(2019)Baevski, Edunov, Liu, Zettlemoyer, and
  Auli]{baevski2019cloze}
Alexei Baevski, Sergey Edunov, Yinhan Liu, Luke Zettlemoyer, and Michael Auli.
\newblock Cloze-driven pretraining of self-attention networks.
\newblock \emph{arXiv}, abs/1903.07785, 2019.

\bibitem[Chen et~al.(2018)Chen, Shen, Huang, Lee, and Lee]{chen2018unsup}
Yi{-}Chen Chen, Chia{-}Hao Shen, Sung{-}Feng Huang, Hung{-}yi Lee, and
  Lin{-}Shan Lee.
\newblock Almost-unsupervised speech recognition with close-to-zero resource
  based on phonetic structures learned from very small unpaired speech and text
  data.
\newblock \emph{arXiv}, abs/1810.12566, 2018.

\bibitem[Chorowski et~al.(2019)Chorowski, Weiss, Bengio, and van~den
  Oord]{chorowski2019unsup}
Jan Chorowski, Ron~J. Weiss, Samy Bengio, and A{\"{a}}ron van~den Oord.
\newblock Unsupervised speech representation learning using wavenet
  autoencoders.
\newblock \emph{arXiv}, abs/1901.08810, 2019.

\bibitem[Chung et~al.(2018)Chung, Weng, Tong, and Glass]{chung2018unsup}
Yu{-}An Chung, Wei{-}Hung Weng, Schrasing Tong, and James~R. Glass.
\newblock Unsupervised cross-modal alignment of speech and text embedding
  spaces.
\newblock \emph{arXiv}, abs/1805.07467, 2018.

\bibitem[Collobert et~al.(2011)Collobert, Weston, Bottou, Karlen, Kavukcuoglu,
  and Kuksa]{collobert2011jmlr}
Ronan Collobert, Jason Weston, L{\'e}on Bottou, Michael Karlen, Koray
  Kavukcuoglu, and Pavel Kuksa.
\newblock Natural language processing (almost) from scratch.
\newblock \emph{JMLR}, 2011.

\bibitem[Collobert et~al.(2016)Collobert, Puhrsch, and
  Synnaeve]{collobert2016wav2letter}
Ronan Collobert, Christian Puhrsch, and Gabriel Synnaeve.
\newblock Wav2letter: an end-to-end convnet-based speech recognition system.
\newblock \emph{arXiv}, abs/1609.03193, 2016.

\bibitem[Collobert et~al.(2019)Collobert, Hannun, and
  Synnaeve]{collobert2018diffbeam}
Ronan Collobert, Awni Hannun, and Gabriel Synnaeve.
\newblock A fully differentiable beam search decoder.
\newblock \emph{arXiv}, abs/1902.06022, 2019.

\bibitem[Dauphin et~al.(2017)Dauphin, Fan, Auli, and
  Grangier]{dauphin2016convlm}
Yann~N Dauphin, Angela Fan, Michael Auli, and David Grangier.
\newblock Language modeling with gated convolutional networks.
\newblock In \emph{Proc. of ICML}, 2017.

\bibitem[Deng et~al.(2009)Deng, Dong, Socher, Li, Li, and
  Fei-Fei]{deng2009imagenet}
Jia Deng, Wei Dong, Richard Socher, Li-Jia Li, Kai Li, and Li~Fei-Fei.
\newblock Imagenet: A large-scale hierarchical image database.
\newblock In \emph{Proc. of CVPR}, 2009.

\bibitem[Devlin et~al.(2018)Devlin, Chang, Lee, and Toutanova]{devlin2018bert}
Jacob Devlin, Ming-Wei Chang, Kenton Lee, and Kristina Toutanova.
\newblock Bert: Pre-training of deep bidirectional transformers for language
  understanding.
\newblock \emph{arXiv}, abs/1810.04805, 2018.

\bibitem[Doersch et~al.(2015)Doersch, Gupta, and Efros]{doersch2015unsup}
Carl Doersch, Abhinav Gupta, and Alexei~A. Efros.
\newblock Unsupervised visual representation learning by context prediction.
\newblock In \emph{Proc. of ICCV}, 2015.

\bibitem[Edunov et~al.(2019)Edunov, Baevski, and Auli]{edunov2019pretrain}
Sergey Edunov, Alexei Baevski, and Michael Auli.
\newblock Pre-trained language model representations for language generation.
\newblock In \emph{Proc. of NAACL}, 2019.

\bibitem[Garofolo et~al.(1993{\natexlab{a}})Garofolo, Graff, Paul, and
  Pallett]{garofolo1993wsj}
John~S. Garofolo, David Graff, Doug Paul, and David~S. Pallett.
\newblock {CSR-I (WSJ0) Complete LDC93S6A. Web Download}.
\newblock \emph{Linguistic Data Consortium}, 1993{\natexlab{a}}.

\bibitem[Garofolo et~al.(1993{\natexlab{b}})Garofolo, Lamel, Fisher, Fiscus,
  Pallett, and Dahlgren]{garofolo1993timit}
John~S. Garofolo, Lori~F. Lamel, William~M. Fisher, Jonathon~G. Fiscus,
  David~S. Pallett, and Nancy~L. Dahlgren.
\newblock {The DARPA TIMIT Acoustic-Phonetic Continuous Speech Corpus CDROM}.
\newblock \emph{Linguistic Data Consortium}, 1993{\natexlab{b}}.

\bibitem[Ghahremani et~al.(2017)Ghahremani, Manohar, Hadian, Povey, and
  Khudanpur]{ghahremani2017asru}
Pegah Ghahremani, Vimal Manohar, Hossein Hadian, Daniel Povey, and Sanjeev
  Khudanpur.
\newblock Investigation of transfer learning for asr using lf-mmi trained
  neural networks.
\newblock In \emph{Proc. of ASRU}, 2017.

\bibitem[Hadian et~al.(2018)Hadian, Sameti1, Povey, and
  Khudanpur]{hadian2018interspeech}
Hossein Hadian, Hossein Sameti1, Daniel Povey, and Sanjeev Khudanpur.
\newblock End-to-end speech recognition using lattice-free mmi.
\newblock In \emph{Proc. of Interspeech}, 2018.

\bibitem[Heafield et~al.(2013)Heafield, Pouzyrevsky, Clark, and
  Koehn]{heafield2013kenlm}
Kenneth Heafield, Ivan Pouzyrevsky, Jonathan~H. Clark, and Philipp Koehn.
\newblock Scalable modified {Kneser-Ney} language model estimation.
\newblock In \emph{Proc. of ACL}, 2013.

\bibitem[H{\'{e}}naff et~al.(2019)H{\'{e}}naff, Razavi, Doersch, Eslami, and
  van~den Oord]{oord2019cpc}
Olivier~J. H{\'{e}}naff, Ali Razavi, Carl Doersch, S.~M.~Ali Eslami, and
  A{\"{a}}ron van~den Oord.
\newblock Data-efficient image recognition with contrastive predictive coding.
\newblock \emph{arXiv}, abs/1905.09272, 2019.

\bibitem[Kamper et~al.(2017)Kamper, Jansen, and Goldwater]{kamper2017seg}
Herman Kamper, Aren Jansen, and Sharon Goldwater.
\newblock A segmental framework for fully-unsupervised large-vocabulary speech
  recognition.
\newblock \emph{Comput. Speech Lang.}, 46\penalty0 (C), November 2017.

\bibitem[Kingma \& Ba(2015)Kingma and Ba]{kingma2015adam}
Diederik~P. Kingma and Jimmy Ba.
\newblock {Adam: A Method for Stochastic Optimization}.
\newblock In \emph{Proc. of ICLR}, 2015.

\bibitem[Kunze et~al.(2017)Kunze, Kirsch, Kurenkov, Krug, Johannsmeier, and
  Stober]{kunze2017transfer}
Julius Kunze, Lous Kirsch, Ilia Kurenkov, Andreas Krug, Jens Johannsmeier, and
  Sebastian Stober.
\newblock Transfer learning for speech recognition on a budget.
\newblock In \emph{Proc. of Workshop on Representation Learning for NLP}, 2017.

\bibitem[Lample \& Conneau(2019)Lample and Conneau]{lample2019cross}
Guillaume Lample and Alexis Conneau.
\newblock Cross-lingual language model pretraining.
\newblock \emph{arXiv}, abs/1901.07291, 2019.

\bibitem[Lian et~al.(2018)Lian, Li, Tao, and Huang]{lian2018emotion}
Zheng Lian, Ya~Li, Jianhua Tao, and Jian Huang.
\newblock Improving speech emotion recognition via transformer-based predictive
  coding through transfer learning.
\newblock \emph{arXiv}, abs/1811.07691, 2018.

\bibitem[Likhomanenko et~al.(2019)Likhomanenko, Synnaeve, and
  Collobert]{likhomanenko2019convlm}
Tatiana Likhomanenko, Gabriel Synnaeve, and Ronan Collobert.
\newblock Who needs words? lexicon-free speech recognition.
\newblock In \emph{Proc. of Interspeech}, 2019.

\bibitem[Lin et~al.(2014)Lin, Maire, Belongie, Hays, Perona, Ramanan,
  Doll{\'a}r, and Zitnick]{lin2014microsoft}
Tsung-Yi Lin, Michael Maire, Serge Belongie, James Hays, Pietro Perona, Deva
  Ramanan, Piotr Doll{\'a}r, and C~Lawrence Zitnick.
\newblock Microsoft coco: Common objects in context.
\newblock In \emph{Proc. of ECCV}, 2014.

\bibitem[Loshchilov \& Hutter(2016)Loshchilov and Hutter]{loshchilov2016cosine}
Ilya Loshchilov and Frank Hutter.
\newblock {SGDR:} stochastic gradient descent with restarts.
\newblock \emph{arXiv}, abs/1608.03983, 2016.

\bibitem[Mikolov et~al.(2013)Mikolov, Sutskever, Chen, Corrado, and
  Dean]{mikolov2013word2vec}
Tomas Mikolov, Ilya Sutskever, Kai Chen, Greg~S Corrado, and Jeff Dean.
\newblock Distributed representations of words and phrases and their
  compositionality.
\newblock In \emph{Proc. of NIPS}, 2013.

\bibitem[Ott et~al.(2019)Ott, Edunov, Baevski, Fan, Gross, Ng, Grangier, and
  Auli]{ott2019fairseq}
Myle Ott, Sergey Edunov, Alexei Baevski, Angela Fan, Sam Gross, Nathan Ng,
  David Grangier, and Michael Auli.
\newblock fairseq: A fast, extensible toolkit for sequence modeling.
\newblock In \emph{Proc. of NAACL System Demonstrations}, 2019.

\bibitem[Panayotov et~al.(2015)Panayotov, Chen, Povey, and
  Khudanpur]{panayotov2015librispeech}
Vassil Panayotov, Guoguo Chen, Daniel Povey, and Sanjeev Khudanpur.
\newblock Librispeech: an asr corpus based on public domain audio books.
\newblock In \emph{Proc. of ICASSP}, pp.\  5206--5210. IEEE, 2015.

\bibitem[Pavllo et~al.(2019)Pavllo, Feichtenhofer, Grangier, and
  Auli]{pavllo2019cvpr}
Dario Pavllo, Christoph Feichtenhofer, David Grangier, and Michael Auli.
\newblock 3d human pose estimation in video with temporal convolutions and
  semi-supervised training.
\newblock In \emph{Proc. of CVPR}, 2019.

\bibitem[Pratap et~al.(2018)Pratap, Hannun, Xu, Cai, Kahn, Synnaeve,
  Liptchinsky, and Collobert]{pratap2018w2l}
Vineel Pratap, Awni Hannun, Qiantong Xu, Jeff Cai, Jacob Kahn, Gabriel
  Synnaeve, Vitaliy Liptchinsky, and Ronan Collobert.
\newblock wav2letter++: The fastest open-source speech recognition system.
\newblock \emph{arXiv}, abs/1812.07625, 2018.

\bibitem[Radford et~al.(2018)Radford, Narasimhan, Salimans, and
  Sutskever]{radford2018unsup}
Alec Radford, Karthik Narasimhan, Tim Salimans, and Ilya Sutskever.
\newblock Improving language understanding by generative pre-training.
\newblock
  \url{https://s3-us-west-2.amazonaws.com/openai-assets/research-covers/language-unsupervised/language_understanding_paper.pdf},
  2018.

\bibitem[Ravanelli \& Bengio(2018)Ravanelli and Bengio]{ravanelli2018mutual}
Mirco Ravanelli and Yoshua Bengio.
\newblock Learning speaker representations with mutual information.
\newblock \emph{arXiv}, abs/1812.00271, 2018.

\bibitem[Ravanelli et~al.(2018)Ravanelli, Brakel, Omologo, and
  Bengio]{ravanelli2018light}
Mirco Ravanelli, Philemon Brakel, Maurizio Omologo, and Yoshua Bengio.
\newblock Light gated recurrent units for speech recognition.
\newblock \emph{IEEE Transactions on Emerging Topics in Computational
  Intelligence}, 2\penalty0 (2):\penalty0 92--102, 2018.

\bibitem[Synnaeve \& Dupoux(2016{\natexlab{a}})Synnaeve and
  Dupoux]{synnaeve2016coherence}
Gabriel Synnaeve and Emmanuel Dupoux.
\newblock A temporal coherence loss function for learning unsupervised acoustic
  embeddings.
\newblock In \emph{Proc. of SLTU}, 2016{\natexlab{a}}.

\bibitem[Synnaeve \& Dupoux(2016{\natexlab{b}})Synnaeve and
  Dupoux]{synnaeve2016temporal}
Gabriel Synnaeve and Emmanuel Dupoux.
\newblock A temporal coherence loss function for learning unsupervised acoustic
  embeddings.
\newblock \emph{Procedia Computer Science}, 81:\penalty0 95--100,
  2016{\natexlab{b}}.

\bibitem[van~den Oord et~al.(2018)van~den Oord, Li, and Vinyals]{oord2018cpc}
A{\"{a}}ron van~den Oord, Yazhe Li, and Oriol Vinyals.
\newblock Representation learning with contrastive predictive coding.
\newblock \emph{arXiv}, abs/1807.03748, 2018.

\bibitem[Vinyals et~al.(2016)Vinyals, Toshev, Bengio, and
  Erhan]{vinyals2016show}
Oriol Vinyals, Alexander Toshev, Samy Bengio, and Dumitru Erhan.
\newblock Show and tell: Lessons learned from the 2015 {MS\,COCO} image
  captioning challenge.
\newblock \emph{arXiv}, abs/1609.06647, 2016.

\bibitem[Woodland et~al.(1994)Woodland, Odell, Valtchev, and
  Young]{woodland1994large}
Philip~C Woodland, Julian~J Odell, Valtcho Valtchev, and Steve~J Young.
\newblock Large vocabulary continuous speech recognition using htk.
\newblock In \emph{Proc. of ICASSP}, 1994.

\bibitem[Wu \& He(2018)Wu and He]{wu2018gn}
Yuxin Wu and Kaiming He.
\newblock Group normalization.
\newblock \emph{arXiv}, abs/1803.08494, 2018.

\bibitem[Zeghidour et~al.(2018{\natexlab{a}})Zeghidour, Usunier, Kokkinos,
  Schaiz, Synnaeve, and Dupoux]{zeghidour2018filters}
Neil Zeghidour, Nicolas Usunier, Iasonas Kokkinos, Thomas Schaiz, Gabriel
  Synnaeve, and Emmanuel Dupoux.
\newblock Learning filterbanks from raw speech for phone recognition.
\newblock In \emph{Proc. of (ICASSP)}, 2018{\natexlab{a}}.

\bibitem[Zeghidour et~al.(2018{\natexlab{b}})Zeghidour, Xu, Liptchinsky,
  Usunier, Synnaeve, and Collobert]{zeghidour2018w2l}
Neil Zeghidour, Qiantong Xu, Vitaliy Liptchinsky, Nicolas Usunier, Gabriel
  Synnaeve, and Ronan Collobert.
\newblock Fully convolutional speech recognition.
\newblock \emph{arXiv}, abs/1812.06864, 2018{\natexlab{b}}.

\end{thebibliography}

\end{document}